\documentclass{article}

\usepackage{microtype}
\usepackage{graphicx}
\usepackage{subfigure}
\usepackage{booktabs} 

\usepackage{hyperref}

\usepackage[table, svgnames, dvipsnames]{xcolor}
\usepackage{makecell, cellspace, caption}
\usepackage{wrapfig}

\usepackage[accepted]{icml2024}

\usepackage{amsmath}
\usepackage{amssymb}
\usepackage{mathtools}
\usepackage{amsthm}
\usepackage{bm}
\usepackage{multirow}
\usepackage{stmaryrd}
\usepackage{bbm}
\usepackage{pifont}

\usepackage[capitalize,noabbrev]{cleveref}

\theoremstyle{plain}
\newtheorem{theorem}{Theorem}[section]
\newtheorem{proposition}[theorem]{Proposition}

\theoremstyle{definition}

\theoremstyle{remark}

\usepackage[textsize=tiny]{todonotes}

\icmltitlerunning{Understanding the Reasoning Ability of Language Models From the Perspective of Reasoning Paths Aggregation}

\begin{document}

\twocolumn[
\icmltitle{Understanding the Reasoning Ability of Language Models \\ From the Perspective of Reasoning Paths Aggregation}



\icmlsetsymbol{equal}{*}

\begin{icmlauthorlist}
\icmlauthor{Xinyi Wang}{ucsb}
\icmlauthor{Alfonso Amayuelas}{ucsb}
\icmlauthor{Kexun Zhang}{cmu}
\icmlauthor{Liangming Pan}{ucsb}
\icmlauthor{Wenhu Chen}{uwaterloo}
\icmlauthor{William Yang Wang}{ucsb}
\end{icmlauthorlist}

\icmlaffiliation{ucsb}{Department of Computer Science, University of California, Santa Barbara}
\icmlaffiliation{cmu}{Language Technologies Institute, Carnegie Mellon University}
\icmlaffiliation{uwaterloo}{Cheriton School of Computer Science, University of Waterloo}

\icmlcorrespondingauthor{Xinyi Wang}{xinyi\_wang@ucsb.edu}

\icmlkeywords{Large language models}

\vskip 0.3in
]



\printAffiliationsAndNotice{}  

\begin{abstract}
Pre-trained language models (LMs) are able to perform complex reasoning without explicit fine-tuning. To understand how pre-training with a next-token prediction objective contributes to the emergence of such reasoning capability, we propose that we can view an LM as deriving new conclusions by aggregating indirect reasoning paths seen at pre-training time. We found this perspective effective in two important cases of reasoning: logic reasoning with knowledge graphs (KGs) and chain-of-thought (CoT) reasoning. More specifically, we formalize the reasoning paths as random walk paths on the knowledge/reasoning graphs. Analyses of learned LM distributions suggest that a weighted sum of relevant random walk path probabilities is a reasonable way to explain how LMs reason. Experiments and analysis on multiple KG and CoT datasets reveal the effect of training on random walk paths and suggest that augmenting unlabeled random walk reasoning paths can improve real-world multi-step reasoning performance.
\footnote{We open source our code at \url{https://github.com/WANGXinyiLinda/LM_random_walk}}.
\end{abstract}

\section{Introduction} \label{sec:intro}
\begin{figure}[tb]
    \centering
    \small
    \includegraphics[width=0.49\textwidth]{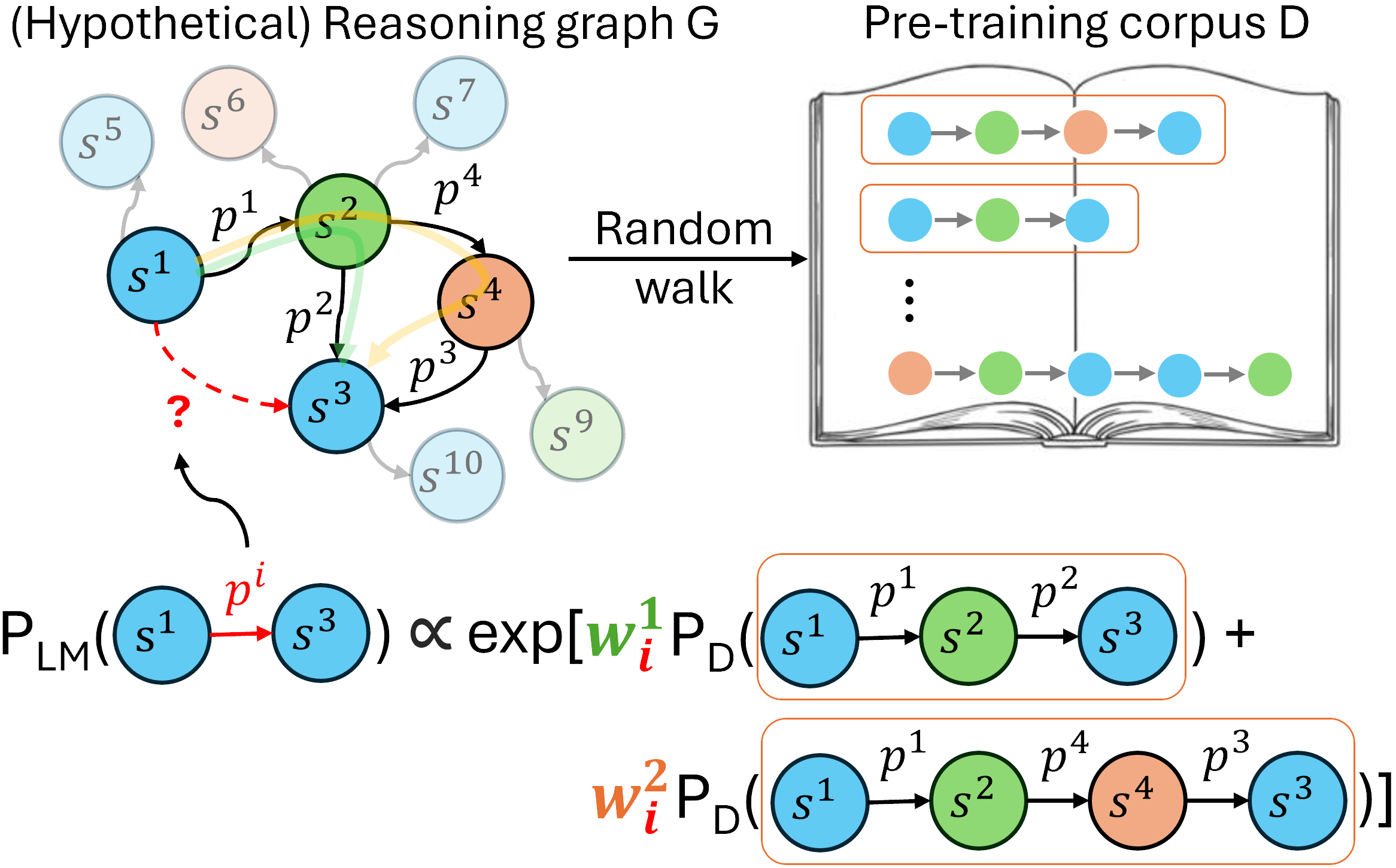}
    \vspace{-1.5em}
    \caption{We hypothesize that the pre-training corpus can be viewed as generated from random walks on a reasoning graph over world knowledge/concepts. With each node $s_i$ representing concepts, $p_j$ can be viewed as arguments that connect them. Then we hypothesize that a language model (LM) training on such a corpus can be viewed as reasoning by a weighted aggregation of random walk paths that connect the entities in interest. $P_{\text{LM}}$ denote the LM distribution while $P_D$ denotes the random walk probability from the pre-training corpus. $w_i^1$ denotes the weight assigned to the first random walk path by the LM for argument $p_i$, and $w_i^2$ denotes the weight assigned to the second random walk path.}
    \label{fig:overview}
    \vspace{-10pt}
\end{figure}

Recently, pre-trained large language models (LLMs) \citep{touvron2023llama,touvron2023llama2,brown2020language} have demonstrated remarkable capabilities in performing intricate reasoning tasks \cite{kojima2022large}. These tasks include problem-solving with world knowledge \cite{hendrycks2020measuring, suzgun2022challenging}, logical reasoning \cite{pan-etal-2023-logic}, and solving mathematical problems \cite{cobbe2021training, hendrycks2021measuring}. These models are typically not explicitly fine-tuned to solve these tasks. Recent research \cite{jain2023mechanistically} also suggests that the supervised fine-tuning process following pre-training only learns a wrapper on top of the already existing model capabilities, instead of learning new ones. It is intriguing to understand how next-token prediction pre-training contributes to the emergence of such reasoning capability. A better understanding of this matter can also inspire new pre-training/fine-tuning techniques to improve these important abilities of LLMs.

It is well-known that LLMs acquire emergent abilities through extensive pre-training \cite{wei2022emergent}. 
In this paper, we focus on elucidating the emergence of reasoning ability — the capacity to draw novel conclusions from existing knowledge, which has been less studied. Many recent works also attempt to understand this phenomenon. Some works focus on understanding Transformers' reasoning capability by construction \cite{liu2023transformers, chi2023transformer, feng2023towards}. Others try to provide post hoc mechanistic explanations \cite{geiger2021causal, wu2023interpretability, hanna2023does} or understanding inference time in-context learning reasoning \cite{li2023dissecting, razeghi2022impact, wang-etal-2023-towards}. Our study is more relevant to the line of work analyzing the contribution of pre-training data to LM reasoning \cite{bi2023program, chen2023program, xiao2023conditions, zhou2023algorithms, ramesh2023capable}. 

In contrast to these works, we adopt a Bayesian view and try to understand why next-token-prediction pre-training can unlock LMs' reasoning ability. More specifically, we hypothesize that LMs can aggregate the indirect reasoning paths seen at pre-training time, through the next-token-prediction training objective. In a real-world scenario, the reasoning path can be a piece of text argument connecting two concepts. We hypothesize that, at inference time, this enables an LM to jump from one concept to another during its reasoning process, which could be verbalized by generating chain-of-thought (CoT) solutions \cite{wei2022chain}, or silent without generating outputs. 

\citet{prystawski2023think} propose a different hypothesis that localized structure on dependencies between variables in training data is important for LM reasoning, especially CoT reasoning. Our hypothesis implies a similar property of the pre-training data: when two concepts are related by a reasoning path, they are highly likely to cooccur in the data and thus form a graph-like localized structure. One drawback of \citet{prystawski2023think}'s work is that their experiments equate reasoning to conditional probability estimation of boolean variables with intermediate variables, which can be considered overly simplified compared to real-world reasoning processes. In our paper, we aim to produce a more realistic analysis of the effect of training data by closely examining two predominant types of reasoning: logical reasoning and mathematical reasoning. In these two reasoning scenarios, we first construct unsupervised random walk paths, which are used to (continually) pre-train the LM with next-token loss. Then we adopt the pre-trained LM to perform reasoning tasks on unseen examples. 

For logical reasoning, we analyze a straightforward yet general reasoning scenario: reasoning over knowledge graphs. A knowledge graph (KG) stores facts in the form of triples $(e_1, r, e_2)$, where $e_1$ and $e_2$ represent entities connected by the relationship $r$. KGs can be incomplete, lacking certain relations between existing entities. These missing relations can typically be inferred from the known triples stored in the KG by employing logical rules. 
For instance, the relation (A, \texttt{isGrandChildof}, C) can be derived from the triples (A, \texttt{isSonOf}, B) and (B, \texttt{isSonOf}, C). We formalize a reasoning path as a \textbf{random walk path} on the KG, which enables us to accurately compute its probability. We show that an LM pre-trained from scratch on random walk paths generated from a given KG can accurately deduce missing relation connections. We also analyze the KL divergence between LM output distributions and weighted/unweighted sums of random walk path probabilities, which are variances of the classic path ranking algorithm (PRA) \citep{lao-etal-2011-random}. Our analysis suggests that the LM distribution shares many similarities with aggregating the probabilities of possible random walk paths in a logical-rule-aware manner, and is usually superior to them. 

For mathematical reasoning, we focus on a more complex case of reasoning: solving math word problems (MWPs). Since it is very challenging to pre-train an LM from scratch to perform well on MWPs, which require both math deduction and language understanding, we propose to continue training on a pre-trained base LM. Based on the insights obtained from the KG reasoning analysis, We propose to create \textbf{random walk reasoning paths} from existing CoT training data, and test the effectiveness of next-token-prediction training on these unlabeled reasoning paths. More specifically, we construct a reasoning graph by regarding the reasoning state at each CoT step as the graph node. Then we reorder and reconnect the existing CoT steps to form the random walk paths on the graph. 
Experiment results on three MWP datasets, GSM8K \cite{cobbe2021training}, AQUA \cite{ling-etal-2017-program}, SVAMP \cite{patel-etal-2021-nlp}, show consistent improvement compared to vanilla supervised fine-tuning, and a similar effect of random walk path length as in the KG reasoning case is observed.


Our findings can be summarized as follows: (a) We show in both reasoning scenarios that our weighted random walk reasoning paths aggregation hypothesis is one (of many) valid ways to explain how LMs may gain their reasoning ability; (b) We show that LMs can utilize unlabeled reasoning paths highly efficiently and show the potential of incorporating the random walk idea to real-world (continue) pre-training. 
\section{Logical Reasoning}

We first analyze a well-controlled case of logic reasoning, knowledge graph (KG) reasoning, by pre-training a small Transformer over random walk paths from KGs. The KL divergence between aggregated random walk path probabilities and LM distribution shows that LM is very close to a weighted aggregation. We also show that KL divergence reflects how LMs assign weights to logical rules. We find that there is usually an optimal random walk path length for training LMs. These observations support our reasoning paths aggregation hypothesis.

\subsection{Problem setting}


Consider a knowledge graph $\mathcal{G}=\{(e^i_1, r^i, e_2^i)\}_{i=1}^N$ consisting of $N$ triples, such that the head entity $e^i_1$ and tail entity $e_2^i$ are related by $r^i$ for all $i$. Let $\mathcal{R}$ denote the set of all possible relations and $\mathcal{E}$ denote the set of all entities. Our goal is to predict a set of unseen triples $\mathcal{T} = \{(e^j_1, r^j, e_2^j)\}_{j=1}^m$, $e^j_1, e_2^j \in \mathcal{E}$, $r^j \in \mathcal{R}$, by training a Transformer based generative language model (LM) from scratch on the given knowledge graph $\mathcal{G}$. To translate a triple into a sentence (i.e. a sequence of tokens), We add each entity $e^i$ and relation $r^i$ as a new token (\texttt{<e\_i>} and \texttt{<r\_i>}) to the Transformer's vocabulary and translate each triple into a three-token sentence ``\texttt{<e\_i> <r\_j> <e\_k>.}". In this way, we avoid using any natural language thus no semantic meaning of the entity or relation name will affect the LM prediction.

\subsection{Language Model Pre-training}

We construct the training data by performing random walks on the given KG $\mathcal{G}$. More specifically, we randomly sample a start entity $e \sim U(\mathcal{E})$, where $U(\cdot)$ denotes the uniform distribution. Then we perform a random walk on $\mathcal{G}$ from $e$ by sampling the next node with $e' \sim U(C(e))$, and stop at a maximum path length $L_{max}$.

Then we translate each triple into a sentence and concatenate all the sentences in the sampled random walk path to become a paragraph. The paragraphs are then concatenated together and separated by the special end-of-sequence token to form text chunks of the same length. The training loss function is the next-token prediction loss:
\begin{align}
    \mathcal{L}_{LM}(\theta) = \sum_\mathcal{D} \sum_{t=1}^T \log \frac{\exp{(f_\theta(w_{t+1}|w_{1:t}))}}{\sum_{w \in \mathcal{V}} \exp{(f_\theta(w|w_{1:t}))}} 
\end{align}

Here, $\theta$ denotes the LM parameters \footnote{We use a randomly initialized GPT-2 model \cite{radford2019language}.}. $w_i \in \mathcal{V}$ represents a token in the LM vocabulary $\mathcal{V}$, and $w_{1:T}$ is a token sequence in the training data $\mathcal{D}$, where $T$ is the length of a text chunk.

To test the reasoning ability of a pre-trained LM, we format the testing triples as sentence completion tasks. For example, the triple ($e_1$, $r$, $e_2$) will be translated to the prompt ``\texttt{<e\_1>} \texttt{<r>}) ", and let the LM predict the next token, then verify the prediction with the ground truth $e_2$. Note that, here the raw LM output distribution is over all entities and relations. To make the LM distribution more well-defined and simplify the following analysis, we take the LM output logits over all entities and define the LM output distribution as:
\begin{align}
    P_{\text{LM}}(e_2|e_1, r) = \frac{\exp{(f_\theta(e_2|e_1,r))}}{\sum_{e \in \mathcal{E}} \exp{(f_\theta(e|e_1,r))}} 
    \label{eq:lm_dist}
\end{align}

\subsection{Random Walk Paths Aggregation}

Recall that our hypothesis is LM can aggregate the reasoning paths seen at the pre-training time. In the KG setting, we can explicitly define how the reasoning/random walk paths are aggregated. Inspired by the classic path ranking algorithm PRA~\cite{lao-etal-2011-random}, \textbf{we define the aggregation of random walk paths $P_w$} as the exponential of a weighted sum of the probabilities of all appropriate random walk paths connecting the two target entities. More specifically, we are interested in a distribution $P_w(e_2|e_1,r)$ for unseen $(e_1, r, e_2)$ in the form of:
\begin{align}
    \label{eq:w_dist}
    P_w(e_2|e_1,r) = \frac{\exp(S_w(e_2 | e_1, r)/T)}{\sum_{e \in \mathcal{E}} \exp(S_w(e | e_1, r)/T)}
\end{align}
Here $S_w(e_2 | e_1, r)$ is a score/logits of $e_2$. $T>0$ is a temperature to rescale the weighted logits $S_w$ so that it can match the scale of LM logits $f_\theta$ \footnote{In practice, we take $T=0.01$.}, and that $P_w(e_2|e_1,r)$ and $P_{\text{LM}}(e_2|e_1, r)$ are more comparable. The score $S_w(e_2 | e_1, r)$ is defined to be a weighted sum of the probability of following all possible logical rules going from $e_1$ to $e_2$:
\begin{align*}
    S_w(e_2 | e_1, r) = \sum_{h \in \mathcal{H}} w_r(h) P(e_2 | e_1, h)
\end{align*}
Here $\mathcal{H}$ denotes the set of all possible logical rules, and $h\in\mathcal{H}$ is a specific logical rule. $w_r(h)$ is the weight assigned to rule $h$ when inferring relation $r$.
For example, a rule for inferring the \texttt{locatedIn} relation can be $h:$ ($e_1$, \texttt{neighborOf}, $e_3$) $\wedge$ ($e_3$, \texttt{locatedIn}, $e_2$).
Formally, for a target relation $r$, we consider logic rules with conjunctive form. $\forall \{e_i\}_{i=0}^n \subset \mathcal{E}$,
\begin{align*}
    (e_0, r, e_n) \leftarrow (e_0, r_1, e_1) \wedge ... \wedge (e_{n-1}, r_n, e_n)
\end{align*}
where $(e_{i-1}, r_i, e_i) \in \mathcal{G}$. We abbreviate such rule by $h = [r_1, r_2, ... ,r_n]$. We can formalize the set of all possible logic rules by $\mathcal{H} = \{[r_1, r_2, ... ,r_n]|n \geq 1, r_i \in \mathcal{R}\}$.

Then the probability of following a specific logic rule $h \in \mathcal{H}$ between $e_1$ and $e_2$ during the random walk would be the sum of the probability of all possible random walk paths from $e_1$ to $e_2$ following the rule $h = [r_1, r_2, ...,r_n]$:
\begin{align*}
    P(e_n | e_0, h) = \sum_{(e_0, r_1, e_1) ... (e_{n-1}, r_n, e_n) \in \mathcal{P}_h} \prod_{i=1}^n P(e_i | e_{i-1}, r_i)
\end{align*}
where $\mathcal{P}_h$ denotes all paths from the KG following $h$. Following the pre-training data generation, we perform a uniform random walk. i.e. $P(e_i | e_{i-1}, r_i) = 1/|C(e_{i-1})|$. Then the rule probability $P(e_2 | e_1, h)$ can be computed directly from the KG.

To learn the rule weights $w_r$, we first observe that
\begin{align*}
    P_w(e_2|e_1,r) = \frac{P_w(r|e_1, e_2)}{\sum_{e \in \mathcal{E}} P_w(r|e_1, e)},
\end{align*}
if we sample $e_1$ and $e_2$ independently and uniformly. Recall \cref{eq:w_dist}, we can instead model $P_w(r|e_1, e_2) \propto \exp S_w(e_2 | e_1, r)$. We can even further simplify it into a binary classification problem $p_i = P_w(\mathbbm{1}_{r^i=r}|e_1^i, e_2^i)$. Then we can use $w_r$ to parameterize a logistic regression model with a loss function:
\begin{align*}
    \mathcal{L}_r(w) = -\sum_i \left[y_i \ln{p_i} + (1-y_i) \ln{(1-p_i)} \right] + \lambda |w|,
\end{align*}
where $p_i = \frac{\exp{S_w(e_2^i | e_1^i, r)}}{1 + \exp{S_w(e_2^i | e_1^i, r)}}$, and the binary label $y_i = \mathbbm{1}_{r^i=r}$. $\lambda |w|$ is a regularization term, and we can take any appropriate norm on $w$. 

At training time, we sample positive triples with relation $r$ and negative triples with other relations from $\mathcal{G}$ as training data. We search over the graph to compute their probability of being reached by each rule $P(e_2 | e_1, h)$ to compute $p_i$. 

For computation efficiency, we only want to search for a subset of more possible logical reasoning rules $\mathcal{H}_r$ in the test set for each relation $r$, and assign $w_r(h) = 0$ for $h \notin \mathcal{H}_r$. Note that a rule can be infinitely long, so we set a maximum rule length $n \leq N_{max}$.
To obtain $\mathcal{H}_r$, we search over $\mathcal{G}$, and record all paths between any two entities that are connected with the relation $r$, and shorter than $N_{max}$. We then collect the rules that have more than $m$ valid paths. 

A simplified version of $P_w$ would be letting $w_r(h)=1$ for all $h$ and $r$. And \textbf{we define this unweighted aggregation distribution to be $P_{s}$}:
\begin{align}
    \label{eq:s_dist}
    P_{s}(e_2 | e_1, r) = \frac{\exp(\sum_{h \in \mathcal{H}_r} P(e_2 | e_1, h)/T)}{\sum_{e \in \mathcal{E}} \exp(\sum_{h \in \mathcal{H}_r} P(e | e_1, h)/T)}
\end{align}

\begin{figure*}[tb]
    \centering
    \small
    \includegraphics[width=\textwidth]{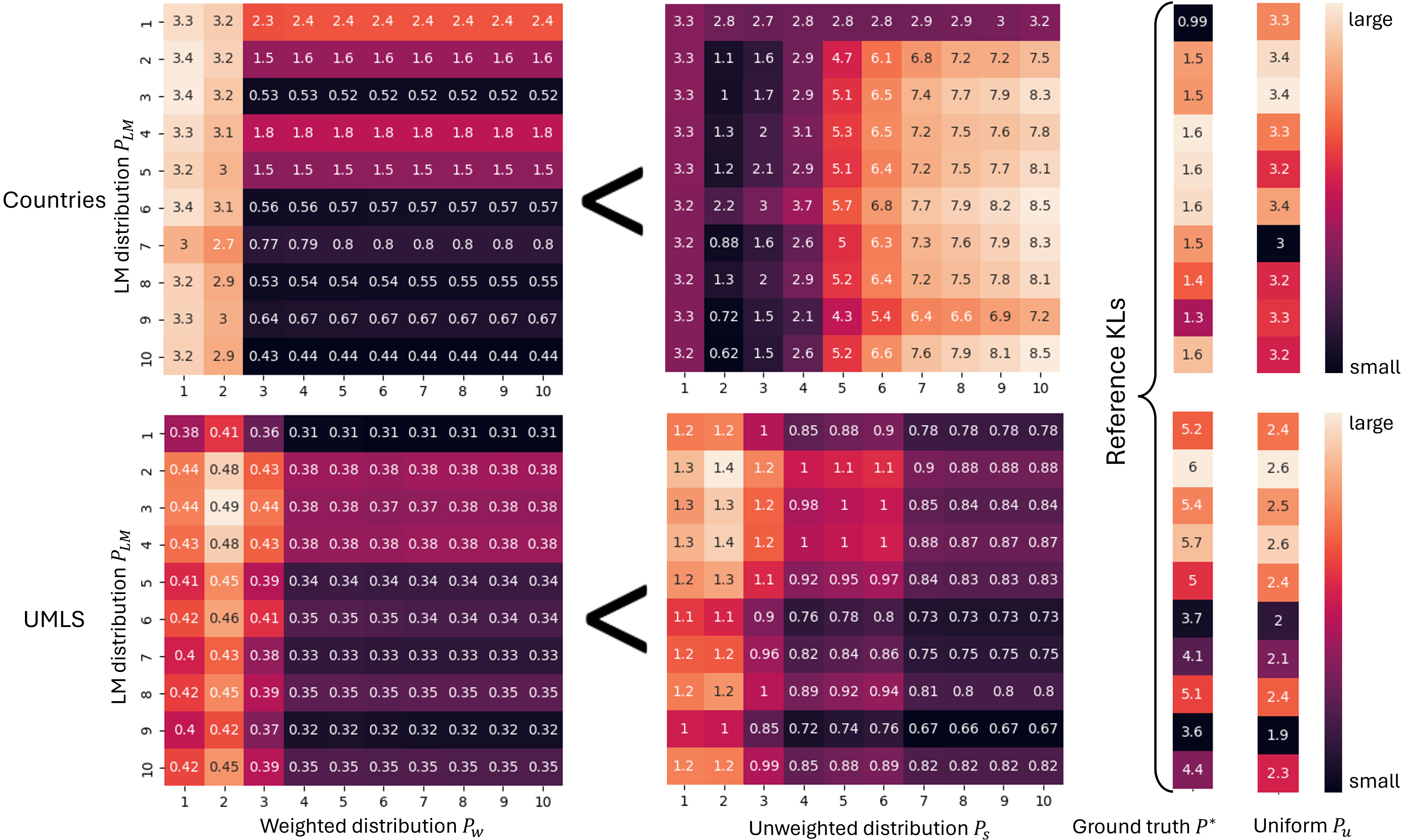}
    \vspace{-10pt}
    \caption{KL divergence between various reference distributions and LM distribution, with different maximum random walk lengths, averaged over Countries (top) and UMLS (bottom) testing set, respectively. The rows correspond to the LM distribution $P_{\text{LM}}(e_2|e_1, r)$ with maximum pre-training random walk path lengths ($L_{max}$) ranging from 1 to 10. From left to right, the columns correspond to the weighted aggregation distribution $P_w(e_2|e_1, r)$ with maximum random walk path lengths  ($N_{max}$) from 1 to 10, the unweighted aggregation distribution $P_s(e_2|e_1, r)$ with maximum random walk path lengths  ($N_{max}$) from 1 to 10, the reference distribution $P^*(e_2|e_1, r)$, and the uniform distribution $P_u(e_2)$, respectively. A \textbf{darker color} represents a \textbf{smaller KL value}, meaning that the two distributions are closer. In general, $KL[P_w, P_{\text{LM}}]$ is always smaller than $KL[P_s, P_{\text{LM}}]$, which implies that LM is learning the difference in rule importance. $KL[P^*, P_{\text{LM}}]$ and $KL[P_u, P_{\text{LM}}]$ serve as anchor points to show the scale of KL values. $KL[P^*, P_{\text{LM}}]$ is generally high because the probability mass concentrates on correct answers, thus it can be very different from the LM distribution. Thus $KL[P^*, P_{\text{LM}}]$ shows how peaky the LM distribution is, and $KL[P_u, P_{\text{LM}}]$ shows how flat the LM distribution is.}
    \label{fig:klds}
    \vspace{-10pt}
\end{figure*}

\begin{figure*}[tb]
    \centering
    \small
    \includegraphics[width=\textwidth]{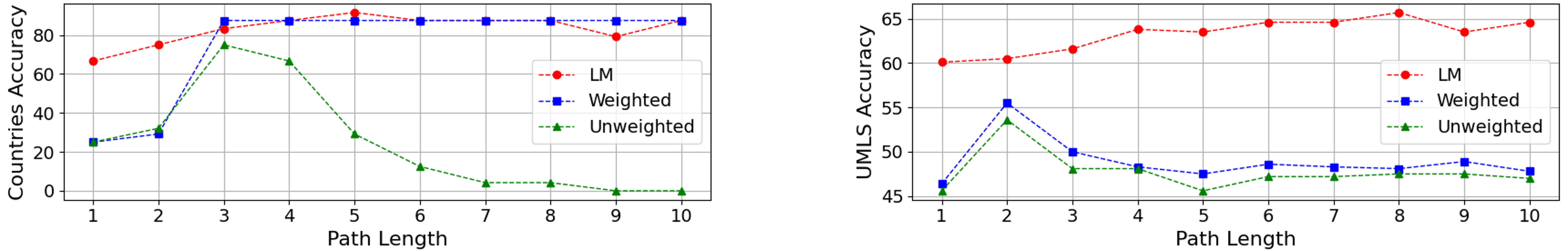}
    \vspace{-10pt}
    \caption{Testing accuracy w.r.t. various maximum pre-training random walk lengths ($1 \leq L_{max} \leq 10$) on Countries (left) and UMLS (right) datasets, respectively. For Countries, the LM ($P_{\text{LM}}$) performance converges to the weighted aggregation ($P_w$) performance, while for UMLS, LM consistently outperforms both weighted ($P_w$) and unweighted ($P_s$) aggregation performance. This is likely because LM ($P_{\text{LM}}$) can learn a better logical rule weighting scheme than weighted aggregation ($P_w$) in more complex KGs.}
    \label{fig:accs}
    \vspace{-10pt}
\end{figure*}

\subsection{KL Divergence and Prediction Accuracy}

To better understand the similarity between LM and the random walk aggregation algorithm as described in the previous section, we propose to compute and analyze the KL divergence between them: $KL[P_w(\mathbf{e}|e_1,r), P_{\text{LM}}(\mathbf{e}|e_1,r)]$, where $\mathbf{e}$ is a random variable taking values in $\mathcal{E}$. To better understand the meaning of the computed KL divergence, we derive an upper bound of it by writing $P_{\text{LM}}(e_2|e_1,r)$ as marginalization over rules:
\begin{align}
    P_{\text{LM}}(e_2|e_1,r) = \sum_{h \in \mathcal{H}} P(e_2 | e_1, h) P_{\text{LM}}(h | e_1, r)
\end{align}
Similarly, we can write   
\begin{align}
    P_w(e_2|e_1,r) = \sum_{h \in \mathcal{H}} P(e_2 | e_1, h) P_w(h | e_1, r)
\end{align}
Then by the Log sum inequality, we can see that the KL divergence of the rule importance is an upper bound of the computed KL divergence \footnote{Proof available in \cref{app:proof}.}:
\begin{proposition}
If LM effectively learned the random walk data distribution through pre-training, we have
\begin{align*}
    &\text{KL}[P_w(\mathbf{e}|e_1,r), P_{\text{LM}}(\mathbf{e}|e_1,r)] \\
    \leq &\text{KL}[P_w(\mathbf{h}|e_1, r), P_{\text{LM}}(\mathbf{h}|e_1,r)]
\end{align*}
\end{proposition}
Here $\mathbf{h}$ is a random variable taking values in $\mathcal{H}$.
This means the KL divergence reflects how LM assigns probabilities to possible logical rules based on the given prompt, which implies how the LM learns to do logical reasoning.

\textbf{KL computation} We compute the KL divergence between the weighted aggregation distribution $P_w(e_2|e_1, r)$ as defined in \cref{eq:w_dist} and the LM distribution $P_{\text{LM}}(e_2|e_1, r)$ as defined in \cref{eq:lm_dist}, abbreviated as $KL[P_w, P_{\text{LM}}]$. We then compare it with the KL divergence between the unweighted aggregation distribution $P_s(e_2|e_1, r)$ as defined in \cref{eq:s_dist} and the LM distribution, abbreviated as $KL[P_s, P_{\text{LM}}]$. To better understand the effect of random walk length, we consider maximum random walk path length ranging from 1 to 10 (i.e. $1 \leq L_{max} \leq 10$ and $1 \leq N_{max} \leq 10$), for computing both the aggregation distribution and the LM distribution. We then compute a pairwise KL between each of them and show the results as a heatmap. To better anchor the computed KL divergence, we also compute the KL divergence $KL[P^*, P_{\text{LM}}]$ between a reference distribution $P^*$ and the LM distribution $P_{\text{LM}}$, and KL divergence $KL[P_u, P_{\text{LM}}]$ between the uniform distribution $P_u$ and the LM distribution $P_{\text{LM}}$. Here \textbf{$P^*$ is uniform over all correct answers}, and \textbf{$P_u$ is uniform over all possible answers}. The described KL divergences for Countries (top) and UMLS (bottom) testing sets are shown in heatmaps in \cref{fig:klds}. More interpretations of these quantities can be found in the caption.

\textbf{Accuracy} We also compute the prediction accuracy using each method and plot it w.r.t to path length ($1 \leq L_{max} \leq 10$). Note that there could be more than one correct answer for a query $(e_1, r)$. We say the prediction is correct as long as it is one of the correct answers. The described testing accuracy for Countries (left) and UMLS (right) is shown in \cref{fig:accs}, where \textbf{LM} is $\arg \max P_{\text{LM}}$, \textbf{Weighted} is $\arg \max P_w$, and \textbf{Unweighted} is $\arg \max P_s$. In general, LM predictor $P_{\text{LM}}$ performs on par/better than weighted aggregation $P_w$, and significantly better than the unweighted aggregation $P_s$. This shows that LM likely learns a better logical rule weighting scheme than $P_w$.

\subsection{Results and Analysis}

\begin{figure}[tb]
    \centering
    \small
    \includegraphics[width=0.4\textwidth]{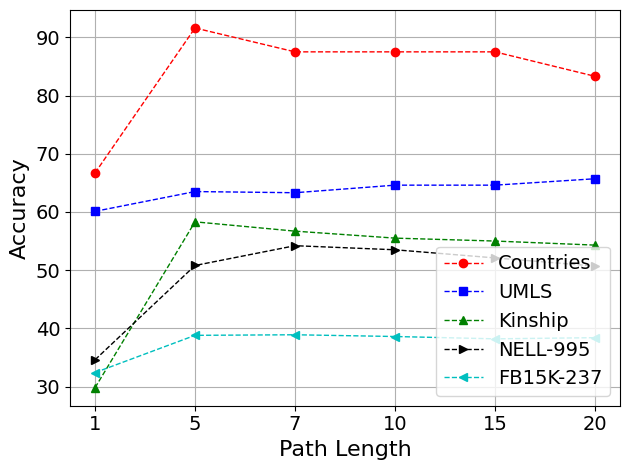}
    \vspace{-5pt}
    \caption{Testing accuracy of LM trained on different random walk path lengths. Each line corresponds to a different KG dataset and thus is not directly comparable. We want to highlight the common trend here that each line peaks at some optimal path length.}
    \vspace{-10pt}
    \label{fig:logic-len-acc}
\end{figure}

We consider five KG datasets in total: Countries \cite{Bouchard2015OnAR}, UMLS \cite{10.1145/1273496.1273551}, Kinship \cite{denham2020}, NELL-995 \cite{xiong2017deeppath}, and FB15K-237 \cite{toutanova-etal-2015-representing} \footnote{More dataset details can be found in \cref{app:data}.}. We take the smallest two for KL divergence analysis for their lower time complexity. We show LM prediction accuracy for all datasets with different pre-training path lengths. 

\textbf{KL divergence with Countries} In \cref{fig:klds} (top), we can see that when the maximum path length for computing the aggregated distribution (columns) is three, there is a sudden drop in $KL[P_w, P_{\text{LM}}]$. This is because the ground truth path length to reach the correct answers in the testing set is three (fixed when constructing the dataset). Both the weighted and unweighted aggregation of random walk paths have low accuracy with path lengths less than three as shown in \cref{fig:accs} (left). The behavior of the path aggregation method is not well-defined at this stage and thus can result in an abnormal KL trend. On the other hand, LM yields a non-trivial accuracy when trained with a path length smaller than three, which shows LM's ability to generalize beyond the pre-training reasoning length. This echoes the findings in \citet{xiao2023conditions, zhou2023algorithms}, that Transformers can generalize to longer sequences than training sequences. 

As shown in \cref{fig:klds} (top left), the weighted aggregation scheme $P_w$ converges to a stable distribution, likely by putting most weights on shorter rules when using long random walk paths. The LM distribution $P_{\text{LM}}$ becomes closer to $P_w$ when the pre-training path length becomes longer. 
On the other hand, $KL[P_s, P_{\text{LM}}]$ stably increases when the path length for $P_s$ becomes larger. This echoes the accuracy trends as shown in \cref{fig:accs} (left). For the countries dataset, since it only has two relations, longer random walk paths introduce more noise than useful information. Thus by increasing the path length the unweighted aggregation scheme $P_s$ becomes less and less effective. Both $P_w$ and $P_{\text{LM}}$ learn to assign a small weight to the long/noisy paths, and thus do not experience an accuracy drop.

\textbf{KL divergence with UMLS} In \cref{fig:klds} (bottom), we can see that when the maximum path length for computing the aggregated distribution (columns) is larger than 3, the weighted aggregation scheme $P_w$ also converges to a stable distribution. To investigate why path length 3 is unique, we find the average path length corresponding to the largest number of valid paths for each relation in the testing set is 3.14. We find the average path length corresponding to the largest weight assigned by $P_w$ when $N_{max}=10$ is 2.75.  This confirms that path length three is likely a good rule length for many relations. However, from \cref{fig:accs} (right), we can see that both weighted ($P_w$) and unweighted ($P_s$) aggregation peaked at path length two instead of three. We believe this is because when the rule length becomes larger (i.e. larger than two), the validity of a rule would be more head entity ($e_1$) dependent. Using only relation-dependent weight $w_r(h)$ as in $P_w$ is likely insufficient. This also explains why LM constantly outperforms both path aggregation methods: LM likely learns a rule importance function that depends both on the head entity and the relation.

Different from the Countries dataset, UMLS' $KL[P_s, P_{\text{LM}}]$ does not increase when the path length for $P_s$ increases. Instead, $KL[P_s, P_{\text{LM}}]$ follows a similar trend as $KL[P_w, P_{\text{LM}}]$, while in general $KL[P_w, P_{\text{LM}}]$ is smaller than $KL[P_s, P_{\text{LM}}]$. Similarly, in \cref{fig:accs} (right), the weighted ($P_w$) and unweighted ($P_s$) aggregation has a similar performance, while $P_w$ is slightly better. This shows that the logical rule weights learned by $P_w$ are similar between different rules, so it has similar effects (KL and accuracy) as the unweighted version $P_s$. The LM also has a flatter distribution, as we can see for UMLS $KL[P^*, P_{\text{LM}}] < KL[P_u, P_{\text{LM}}]$ while for Countries $KL[P^*, P_{\text{LM}}] > KL[P_u, P_{\text{LM}}]$. This is likely because UMLS is more complex than Countries (49 v.s. 2 relations), thus many longer paths and rules are similarly useful for prediction, making the LM distribution flatter.

\textbf{Prediction accuracy v.s. pre-training path length}
We briefly touched on how the pre-training random walk path length $L_{max}$ affects the LM distribution in the analysis above. In general, a longer path length improves the prediction accuracy and decreases $KL[P_w, P_{\text{LM}}]$. This shows that LM can improve the logical rule weight assignment when trained with a longer path length. To further investigate this problem, we pre-train LM on longer random walk path lengths with more KG datasets. 

In \cref{fig:logic-len-acc}, we show the LM prediction accuracy v.s. the maximum pre-training random path length of 1, 5, 7, 10, 15, and 20, trained on five different KG datasets. In general, there is a large performance gain from a path length of 1 to 5. Note that when the path length is equal to one, we randomly sample individual triples from a KG. i.e. There are no reasoning paths in the training data. So it is important to have reasoning paths with a non-trivial length in the pre-training data, to enable the LM's reasoning ability. By extending the maximum length from 10 to 20, we can see that there is a slight drop in the Countries dataset. Similarly, in most datasets, there is a small decrease after an optimal path length. This is likely because a too-long random walk path would contain more noise/unrelated triples for reasoning. i.e. It is less likely to be useful for predicting the head and tail entity relation in a path aggregation sense. On the other hand, we can understand this from a localized data structure perspective \cite{prystawski2023think}: a sufficiently long random walk path makes any two entities similarly possible to appear in the same path, thus hurting the local dependency in the training data.
\section{Chain-of-thoughts Reasoning}

After carefully analyzing the logical reasoning on KGs, we want to apply and verify the obtained insights on a more general and realistic case of reasoning: chain-of-thoughts (CoT) reasoning \cite{wei2022chain} with textual descriptions and step-by-step solutions. We continue training a pre-trained LM with random walk reasoning paths and show that these unlabeled paths consistently benefit CoT reasoning performance across multiple datasets of various tasks, including math reasoning, multihop question answering (QA), and logical deduction. We also observe a similar optimal random walk path length effect as in the KG logical reasoning case, which is associated with the intrinsic reasoning length of different datasets. These results support our reasoning path aggregation hypothesis and imply principles for constructing/augmenting pre-training data.

\subsection{Problem Setting}

Suppose we have a set of training data $\mathcal{D} = \{(x^i, r^i_1, r^i_2, ..., r^i_{n^i}, y^i)\}_i$, where $x^i$ is a question described in the text that needs to be answered. $r^i_1, r^i_2, ..., r^i_{n^i}$ is a chain-of-thought (CoT) solution, where $r^i_j$ is one reasoning step. $y^i$ is the ground truth answer to the question. Since CoT datasets are hard to collect and usually small in size, a model is not likely to generalize to new questions by aggregating reasoning paths over this small set of CoT reasoning paths. Fine-tuning on a pre-trained LM can effectively mitigate this problem since the LM has already seen many other reasoning paths at the pre-training time, but more unlabeled reasoning paths specific to this task would likely improve the testing performance if the path aggregation hypothesis still holds for this task.  


\subsection{Random Walk on Latent Reasoning Graph}

We assume that CoT paths $r^i_1, r^i_2, ..., r^i_{n^i}$ can be regarded as random walk paths sampled from a reasoning graph $\mathcal{G}$, where the nodes are the reasoning states at each step $r^i_j$. The reasoning state can be regarded as a belief that will be updated after each reasoning step. Denote the last hidden state of the pre-trained LM we are going to tune by $f_\theta$. To represent the reasoning state for each step $r^i_j$, we propose to use $f_\theta$ to cumulatively encode all the steps before $r^i_j$, and then average over the sequence dimension, to obtain a fixed dimensional vector $s^i_j$:
\begin{align*}
    s^i_j = \text{avg} \; f_\theta(x^i, r^i_1, r^i_2, ..., r^i_j)
\end{align*}
Assuming similar $s^i_j$'s are sampled from the same node of the latent reasoning graph, we propose to cluster \footnote{In practice we use K-meanings clustering.} similar $s^i_j$'s together to form a node. Suppose we have constructed a graph $\mathcal{G}$ from the CoT dataset $\mathcal{D}$, with nodes $A_1, A_2, ..., A_K$, where $K$ is predefined by the clustering algorithm. 
Each CoT step would be classified into a node. i.e. $r^i_j \in A_m$ for some $m \in [1,k]$. 
Then we can perform random walks on the graph by using the original CoT as links between the nodes as shown in \cref{alg:random_walk}. Then we record the random walk paths produced by \cref{alg:random_walk} and do next-token-prediction training on them for $M$ steps. To make sure the LM can produce a CoT solution and a final answer, we do another $N-M$ step of supervised fine-tuning (SFT) on the original dataset $\mathcal{D}$, for some $N>M$.

\begin{algorithm}[tb]
    \small
   \caption{Random Walk on Latent Graph}
   \label{alg:random_walk}
\begin{algorithmic}
   \STATE {\bfseries Input:} CoT dataset $\mathcal{D}$, latent graph $\mathcal{G}$, maximum path length $L_{max}$.
   \STATE Randomly initialize current node $a = A_k$. Initialize path $p = []$
   \REPEAT
   \STATE Randomly choose a CoT step $r^i_j \in a$.
   \STATE Uniformly sample $m$ from $[1,L]$. 
   \STATE Append $r^i_j, r^i_{j+1}, ..., r^i_{\min\{j+m, n^i\}}$ to path $p$.
   \STATE Suppose $r^i_{\min\{j+m, n^i\}} \in A_l$. Set $a = A_l$.
   \UNTIL{$\text{len}(p) \geq L_{max}$.}
\end{algorithmic}
\end{algorithm}

\begin{table*}[tb]
    \centering
    \small
    \begin{tabular}{cccccccc}
        \toprule
        Model & Method & GSM8K & AQUA & SVAMP & StrategyQA & LogicalDeduction & Avg.\\
        \midrule
        Gemma (2B) & SFT & 24.8 & 31.4 & 56.4 & 54.2 & 50.7 & 43.5\\
        & Ours & \textbf{26.1} & \textbf{33.9} & \textbf{60.3} & \textbf{56.3} & \textbf{51.6} & \textbf{45.6} \\
        \midrule
        Yi (6B) & SFT & 32.2 & 37.0 & 65.8 & 65.8 & 62.2 & 52.6\\
        & Ours & \textbf{33.1} & \textbf{39.8} & \textbf{67.0} & \textbf{70.0} & \textbf{63.3} & \textbf{54.6}\\
        \midrule
        Llama 2 (7B) & SFT & 26.8 & 30.0 & 53.3 & 58.4 & 55.3 & 44.8 \\
         & Ours & \textbf{28.5} & \textbf{34.6} & \textbf{55.8} & \textbf{63.7} & \textbf{56.1} & \textbf{47.7} \\
         \midrule
        Llama 2 (13B) & SFT & 37.1 & 35.0 & 66.4 & 69.5 & 55.7 & 52.7 \\
         & Ours & \textbf{41.2} & \textbf{37.4} & \textbf{69.0} & \textbf{71.2} & \textbf{57.7} & \textbf{55.3} \\
        \bottomrule
    \end{tabular}
    \caption{Testing accuracy of different open source LMs continue pre-trained with our random walk paths and then supervised fine-tuned. The supervised fine-tuning baseline (SFT) is fine-tuned by the same number of total steps. Results are reported on five CoT datasets.}
    \label{tab:main}
    \vspace{-10pt}
\end{table*}

\begin{figure}[tb]
    \centering
    \small
    \includegraphics[width=0.4\textwidth]{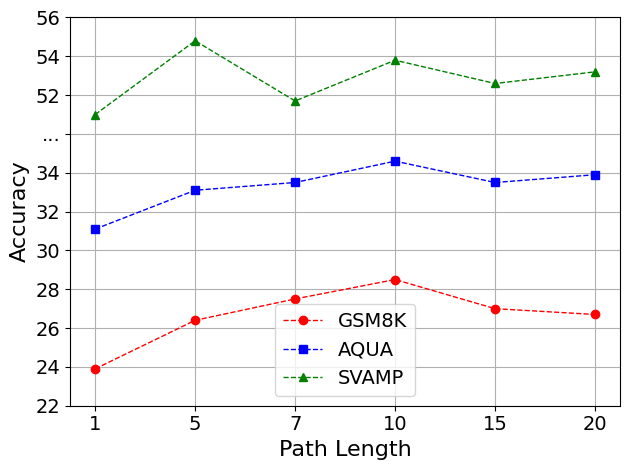}
    \vspace{-10pt}
    \caption{Testing accuracy of continue pre-training with our random walk paths of different length $L_{max}$. Each line corresponds to a different MWP dataset and thus is not directly comparable. We want to highlight the common trend here that each line would peak at some optimal path length range, which is similar to \cref{fig:logic-len-acc}.}
    \label{fig:length}
    \vspace{-10pt}
\end{figure}

\begin{table}[tb]
    \centering
    \small
    \begin{tabular}{cccccc}
        \toprule
        Ablation & & GSM8K & AQUA & SVAMP & Avg.\\
        \midrule
        \#Steps = & 0 & 26.8 & 30.0 & 53.3 & 36.7 \\
        & 200 & 27.5 & 30.1 & 53.6 & 37.1 \\
        & 500 & \textbf{28.5} & \textbf{34.6} & \textbf{55.8} & \textbf{39.6} \\
        & 1000 & 24.9 & 32.3 & 51.6 & 36.3 \\
        \midrule
        \#Nodes = &0 & 26.8 & 30.0 & 53.3 & 36.7 \\
        & 10 & 26.8 & 30.3 & 54.8 & 37.3\\
        & 50 & 26.6 & 29.9 & 54.7 & 37.1 \\
        & 100 & \textbf{28.5} & \textbf{34.6} & \textbf{55.8} & \textbf{39.6} \\
        & 200 & 26.6 & 31.1 & 52.5 & 36.7 \\
        \bottomrule
    \end{tabular}
    \caption{Ablation on the number of random walk training steps $M$ and the number of clusters/nodes $K$.}
    \vspace{-10pt}
    \label{tab:ablation}
\end{table}

\subsection{Experiments}

\textbf{Datasets.} We conduct experiments on three math word problem (MWP) datasets: \textbf{GSM8K} \cite{cobbe2021training}, \textbf{AQUA} \cite{ling-etal-2017-program}, \textbf{SVAMP} \cite{patel-etal-2021-nlp}, a multihop QA dataset \textbf{StrategyQA} \cite{geva2021did}, and a logical deduction dataset \textbf{LogicalDeduction} from the BIG-bench \cite{srivastava2023beyond}. \textbf{GSM8K}, \textbf{AQUA}, and \textbf{SVAMP} are math questions with annotated CoT steps. \textbf{StrategyQA} is annotated with decomposed questions, which we used as the Chain-of-thought (CoT) path of the question. As there is no CoT annotation in \textbf{LogicalDeduction}, we use GPT4 to generate CoTs for the training set, which on average requires 6+ reasoning steps per question. \footnote{More dataset details can be found in \cref{app:data}.}

\textbf{Training} Because of computation limits, we do LoRA \cite{hu2021lora} parameter efficient training in 8 bits with Llama 2 7B and 13B models \cite{touvron2023llama2},  Yi 6B model \cite{young2024yi} and Gemma 2B model \cite{team2024gemma}. If not specified, we default to using the Llama 2 7B model. 

\textbf{Results.} In \cref{tab:main}, we demonstrate the effectiveness of our proposed method against the supervised fine-tuning (SFT) baseline. We train both our method and SFT with $N=2500$ steps in total. The first $M=500$ steps of our method are continually pre-trained on random walk data, and then we do 2000 steps of SFT on the original dataset. Experiment results show that our method can notably improve on math, multihop QA, and logical reasoning. The improvement is especially significant on \textbf{StrategyQA}, likely because of the relative simplicity of the reasoning, as only 3 subquestions per example on average are needed.

Then we investigate the effect of random walk path length $L_{max}$ by plotting accuracy v.s. path lengths. In \cref{fig:length}, we observe that each dataset has a performance peak at a certain random walk length. While both AQUA and GSM8K peak at path length 10, the SVAMP dataset peaks at path length 5. This is likely related to the different intrinsic reasoning lengths for different datasets. The average length of CoTs in AQUA, GSM8, and SVAMP training sets are 4.79, 3.72, and 1.36, respectively. The reasoning length required for SVAMP is significantly shorter than the other two datasets, thus explaining the earlier peaking. As we discussed in the logical reasoning case, a long random walk may introduce more noise than useful information. Note that even the LM performance can drop after the optimal path length, it is always better than training with path length one. i.e. multi-step random walk always helps.

We also do ablation studies on two critical hyperparameters of our method: the number of steps training on random walk paths $M$ and the number of clusters/nodes $K$. In the upper half of \cref{tab:ablation}, we show that the optimal number of training steps $M$ is 500 for all three datasets. Since the generated random walk reasoning paths are not natural within small corpora, e.g. the subject might be suddenly changed from one step to another, training too many steps might make the LM overfit the unwanted artifacts. 
In the lower half of \cref{tab:ablation}, we show that the optimal number of clusters is 100 for all three datasets. Here 0 clusters mean the SFT baseline. Since the datasets we use are small in scale, clustering with a large number of clusters may introduce more noise than useful matchings.
We hypothesize that this may be solved by using a larger dataset and more number of clusters/nodes $K$: in this case, the steps within each node will be more intrinsically similar. This also hints at the potential of our method in the actual pre-training stage: we can view each example in the pre-training corpus as a reasoning path and apply our method.

\begin{figure*}[t]
    \centering
    \includegraphics[width=0.95\textwidth]{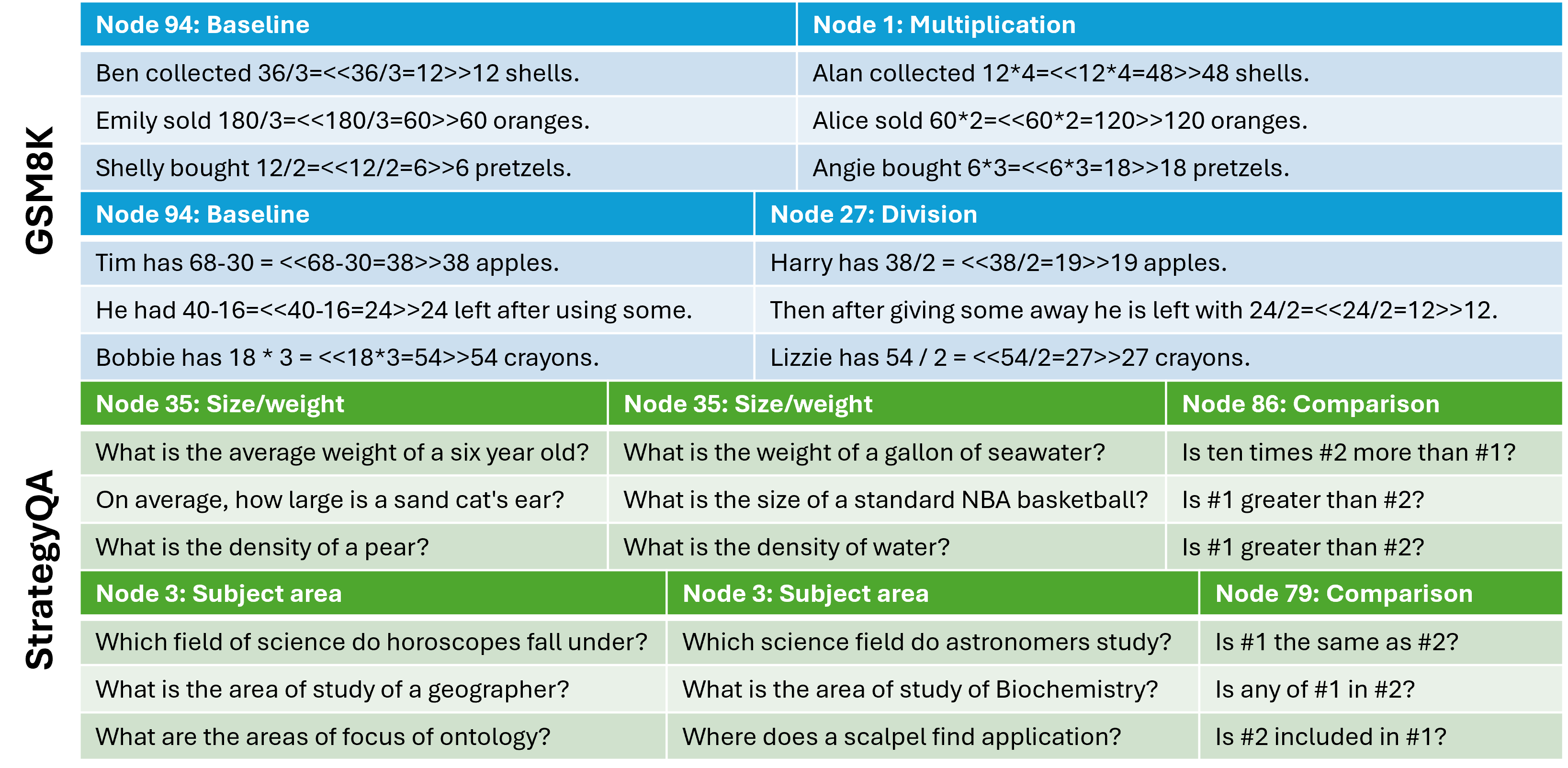}
    \caption{High-frequency node patterns in the training data of \textbf{GSM8K} and \textbf{StrategyQA}, discovered by our constructed latent reasoning graphs, with example CoT solutions belonging to the node pattern.}
    \label{fig:cot_example}
\end{figure*}

\textbf{Latent reasoning graph analysis.} To give a better understanding of the discovered latent reasoning graph, we show some discovered reasoning patterns through the graph in \cref{fig:cot_example}. We show high-frequency node patterns of CoTs in the training set and corresponding CoT examples. We show examples from \textbf{GSM8K} and \textbf{StrategyQA} as they are shorter. With the \textbf{GSM8K} examples, we show our method discovers a 2-step pattern that first computes the baseline quantity and then performs division/multiplication to get the goal quantity based on the question specification. With the \textbf{StrategyQA} examples, we show a 3-step pattern that first decomposes the question into two parallel subquestions of size/weight/subject area, and then uses the third question to compare the answers to the first two questions. \footnote{More latent reasoning graph examples can be found in \cref{app:example}.}

\section{Related Work}

Many recent works have investigated LM's reasoning ability.
\citet{geiger2021causal, wu2023interpretability} ai to find the causal abstraction of an LM. \cite{hanna2023does} tries to find circuit for year-span-prediction. \citet{liu2023transformers, chi2023transformer, feng2023towards} show that CoTs enable fixed-size Transformers to perform certain types of reasoning tasks. 
\citet{li2023dissecting, razeghi2022impact, wang-etal-2023-towards} try to understand inference time in-context CoT reasoning.
Our study is more relevant to the line of work analyzing the contribution of pre-training data to LM reasoning. 
\citet{bi2023program} analyzes how code data affect program-of-thoughts \cite{chen2023program} reasoning ability. \citet{xiao2023conditions, zhou2023algorithms} study how reasoning length generalizes from training data. \citet{ramesh2023capable} studies LMs' compositional generalization ability. Our hypothesis also echos the conclusion of \citet{malach2023auto} that reasoning paths in training data enable supervision on intermediate steps with a next-token-prediction objective. 
\citet{prystawski2023think} propose a different hypothesis that localized structure on dependencies between variables in training data is important for LM reasoning, especially CoT reasoning. Our proposed hypothesis echoes theirs and is shown to be effective on more realistic data and tasks. 
\citet{hou-etal-2023-towards} confirm with attention probing that LMs perform multi-step reasoning internally, which echos our KG logical reasoning results. \footnote{More related work on logical reasoning and math reasoning can be found in \cref{app:related}.}

\section{Conclusion} \label{sec:conclusion}

In conclusion, we aim to understand reasoning abilities in language models (LMs), from the perspective of aggregating reasoning paths from pre-training data. 
The findings shed light on the origins of LLMs' remarkable reasoning capabilities, showcasing the importance of pre-training in acquiring these skills. 
The construction of the pre-training sequence, such as organizing it as "chains" or random walks on the graph, was found to significantly impact the effectiveness of reasoning. 
The study also revealed that LM behavior is similar to reason over known facts by aggregating relevant reasoning paths. 
These insights contribute to our understanding of the underlying mechanisms behind LLMs' reasoning abilities and lead to a potential pre-training data augmentation technique to boost reasoning performance.

\section*{Acknowledgement}
This work was supported by the National Science Foundation award \#2048122. The views expressed are those of the author and do not reflect the official policy or position of the US government.

\section*{Impact Statement}
Understanding the reasoning processes of large language models (LLMs) through the lens of aggregating indirect reasoning paths holds potential implications for identifying and mitigating potential biases within LLMs. By formalizing reasoning as random walk paths on knowledge and reasoning graphs, this approach not only elucidates the mechanisms through which LLMs derive conclusions but also sheds light on data and reasoning paths that contribute to their outputs. This insight is crucial for recognizing biases embedded in the training data or in the reasoning process itself. Recognizing these biases is the first step toward developing more equitable and transparent models. By augmenting models with unbiased, unlabeled random walk reasoning paths, we can potentially reduce the influence of biased reasoning patterns and improve the fairness and reliability of LLMs in real-world applications. This research advances our understanding of LLM reasoning capabilities and their implications for bias, paving the way for more responsible AI development and deployment.

\bibliography{ref}
\bibliographystyle{icml2024}

\newpage
  \appendix

\newtheorem{thmappdef}{Definition}
\renewcommand{\thethmappdef}{A\arabic{thmappdef}}
\newtheorem{thmappasmp}{Assumption}
\renewcommand{\thethmappasmp}{A\arabic{thmappasmp}}
\newtheorem{thmapplem}{Lemma}
\renewcommand{\thethmapplem}{A\arabic{thmapplem}}
\newtheorem{thmappcol}{Corollary}
\renewcommand{\thethmappcol}{A\arabic{thmappcol}}

\newenvironment{thmproof}[1][Proof]{\begin{trivlist}
\item[\hskip \labelsep {\textit{#1.}}]}{\end{trivlist}}
\newenvironment{thmproofsketch}[1][Proof sketch]{\begin{trivlist}
\item[\hskip \labelsep {\textit{#1.}}]}{\end{trivlist}}

\section{Proof}
\label{app:proof}

\begin{proposition}
If LM effectively learned the random walk data distribution through pre-training, we have
\begin{align*}
    \text{KL}[P_w(\mathbf{e}|e_1,r), P_{\text{LM}}(\mathbf{e}|e_1,r)] \leq \text{KL}[P_w(\mathbf{h}|r), P_{\text{LM}}(\mathbf{h}|e_1,r)]
\end{align*}
\end{proposition}

\begin{proof}
    Recall that 
    \begin{align*}
        P_{\text{LM}}(e_2|e_1,r) = \sum_{h \in \mathcal{H}} P(e_2 | e_1, h) P_{\text{LM}}(h | e_1, r)
    \end{align*}
    and
    \begin{align*}
        P_w(e_2|e_1,r) = \sum_{h \in \mathcal{H}} P(e_2 | e_1, h) P_w(h | e_1, r).
    \end{align*}
    By log sum inequality, we have:
    \begin{align*}
        &\text{KL}[P_w(\mathbf{e}|e_1,r), P_{\text{LM}}(\mathbf{e}|e_1,r)] \\
        =& \sum_{e_2\in\mathcal{E}} P_w(e_2|e_1,r) \log \frac{P_w(e_2|e_1,r)}{P_{\text{LM}}(e_2|e_1,r)}\\
        \leq& \sum_{e_2\in\mathcal{E}} \sum_{h \in \mathcal{H}} P(e_2 | e_1, h) P_w(h | e_1, r) \frac{P(e_2 | e_1, h) P_{\text{LM}}(h | e_1, r)}{P(e_2 | e_1, h) P_w(h | e_1, r)} \\
        =& \sum_{e_2\in\mathcal{E}} \sum_{h \in \mathcal{H}} P(e_2 | e_1, h) P_w(h | e_1, r) \frac{P_{\text{LM}}(h | e_1, r)}{P_w(h | e_1, r)} \\
        =& \sum_{h \in \mathcal{H}} (\sum_{e_2\in\mathcal{E}} P(e_2 | e_1, h)) P_w(h | e_1, r) \frac{P_{\text{LM}}(h | e_1, r)}{P_w(h | e_1, r)} \\
        =& \sum_{h \in \mathcal{H}} P_w(h | e_1, r) \frac{P_{\text{LM}}(h | e_1, r)}{P_w(h | e_1, r)} \\
        =& \text{KL}[P_w(\mathbf{h}|r), P_{\text{LM}}(\mathbf{h}|e_1,r)]
    \end{align*}
\end{proof}

\section{Detailed discussion of related work}
\label{app:related}

\textbf{Theory on LM reasoning} Many recent works are investigating LM's reasoning ability. \citet{geiger2021causal, wu2023interpretability} aims to find the causal abstraction of an LM. \cite{hanna2023does} tries to find circuit for year-span-prediction. \citet{liu2023transformers, chi2023transformer, feng2023towards} show that CoTs enable fixed-size Transformers to perform certain types of reasoning tasks. 
\citet{li2023dissecting, razeghi2022impact, wang-etal-2023-towards} try to understand inference time in-context CoT reasoning. Our study is more relevant to the line of work analyzing the contribution of pre-training data to LM reasoning. 
\citet{bi2023program} analyzes how code data affect program-of-thoughts \cite{chen2023program} reasoning ability. \citet{xiao2023conditions, zhou2023algorithms} study how reasoning length generalizes from training data. \citet{ramesh2023capable} studies LMs' compositional generalization ability. Our hypothesis also echos the conclusion of \citet{malach2023auto} that reasoning paths in training data enable supervision on intermediate steps with next-token-prediction objective, and also increase the length complexity, thus reducing time/sample complexity at training time. 
\citet{prystawski2023think} propose a different hypothesis that localized structure on dependencies between variables in training data is important for LM reasoning, especially CoT reasoning. Our proposed hypothesis echoes theirs and can be shown effective on more realistic data and tasks.

\textbf{Logic/knowledge graph reasoning} Existing methods can be divided into three categories: rule-based, GNN-based \cite{GNN}, and LM-based. Markov Logic Network (MLN) \cite{MLN} and path ranking algorithm (PRA) \cite{lao-etal-2011-random} are two classical methods that assign weights to different logical rules. Neural Logic Programming \cite{yang2017differentiable} and RNN-logic \cite{qu2020rnnlogic} are two neural methods that combine the explainability of learned logical rules and the high performance of neural networks. R-GCN \cite{schlichtkrull2018modeling} and NBFNet \cite{NEURIPS2021_f6a673f0} are two GNN-based methods that train a GNN on the KG and use the obtained triple embeddings. These two category methods either rely on random walks to find paths or use random walks to train GNNs. Recently, LM-based methods are shown to be highly effective on not only KG reasoning \cite{misra-etal-2023-triggering}, but more general logical reasoning problems with text descriptions \cite{pan-etal-2023-logic}. 

\textbf{Chain-ot-thought (CoT) reasoning} Recently, LLMs have shown to be highly effective in complex reasoning tasks, like math reasoning \cite{azerbayev2023llemma, yang2023leandojo}. Chain-of-thought (CoT) \cite{wei2022chain} prompting/fine-tuning has been the major way to invoke/improve LLMs' reasoning capabilities. Many variants of CoT prompting have been proposed to improve upon the vanilla CoT prompting \cite{chen2023program, yao2024tree}. 
On the other hand, many works have focused on fine-tuning LLMs on generated high-quality CoT training data \cite{wang2022self, nye2022show, yuan2023scaling}. However, they all rely on the annotated Q-A pairs to generate corresponding paths with LM, which limits the size of augmented data and requires large LMs to do the CoT generation. Our proposed method does not need supervised seed data and thus can be extended to the vast amount of pre-training data. Our method is also lightweight, which only requires a small/medium LM to produce the step embeddings and then do clustering on them.

\section{Experiment Details} \label{app:exp}

\subsection{Datasets}
\label{app:data}

\textbf{Knowledge graph datasets} For KL analysis, we focus on two KGs: Countries \cite{Bouchard2015OnAR} and UMLS \cite{10.1145/1273496.1273551}, as they have a reasonable time complexity to compute the aggregated probabilities for long paths. The Countries \cite{Bouchard2015OnAR} contains two relations (\texttt{locatedIn} and \texttt{neighborOf}) and 227 entities, including countries, regions, and subregions. We use the hardest version (S3) of the Countries. The Unified Medical Language System (UMLS) \cite{10.1145/1273496.1273551} is a more complex KG built from biomedicine knowledge, containing 49 relations and 135 entities. Example entities are diseases and antibiotics, and example relations are treats and diagnoses. 

We add three more datasets for computing the prediction accuracy v.s. different random walk path lengths: Kinship \cite{denham2020}, NELL-995 \cite{xiong2017deeppath}, and FB15K-237 \cite{toutanova-etal-2015-representing}. The Kinship dataset contains 104 entities and 26 kinship relationships among members of the Alyawarra tribe from Central Australia. The NELL-995 dataset contains 75,492 entities and 200 relations, which is built from the Web via an intelligent agent called Never-Ending Language Learner. The FB15K-237 dataset contains 14,505 entities and 237 relations derived from Freebase. We adopt a processed version of these datasets from \citet{das2017go}.

\textbf{Math word problem datasets} We conduct experiments on three math word problem (MWP) datasets: GSM8K \cite{cobbe2021training}, AQUA \cite{ling-etal-2017-program}, SVAMP \cite{patel-etal-2021-nlp}. The Grade School Math dataset (\textbf{GSM8K}) contains 8.5K examples of linguistically diverse grade school math world problems.
The \textbf{AQUA}-RAT dataset contains 100K samples of mathematical problems, along with sequences of human-readable mathematical expressions in natural language.
The \textbf{SVAMP} dataset is a testing set consisting of elementary-level MWPs. The training set is a combination of simpler MWPs: MAWPS \cite{koncel-kedziorski-etal-2016-mawps} and ASDiv-A \cite{miao-etal-2020-diverse} with 3.5k training examples in total.

\textbf{StrategyQA} is annotated with decomposed questions, which we used as the Chain-of-thought (CoT) path of the question. Since the test set labels are not publicly released and the testing set predictions are only allowed to be verified every 7 days, we split the original training set into a new training and testing set.

\subsection{Training Details}

\textbf{Logical reasoning} We train randomly initialized GPT-2 \cite{radford2019language} (124M parameters) with batch size 16 and learning rate 5e-4 using AdamW optimizer \citep{loshchilov2017decoupled} on one 24G Titan GPU.

\textbf{CoT reasoning} We continually (LORA) train all base LLMs with batch size 16 and learning rate 2e-4 using AdamW optimizer \citep{loshchilov2017decoupled} on one 40G A100 GPU.

\subsection{Additional Latent Reasoning Graph Examples}
\label{app:example}

\begin{figure*}[t]
    \centering
    \includegraphics[width=0.95\textwidth]{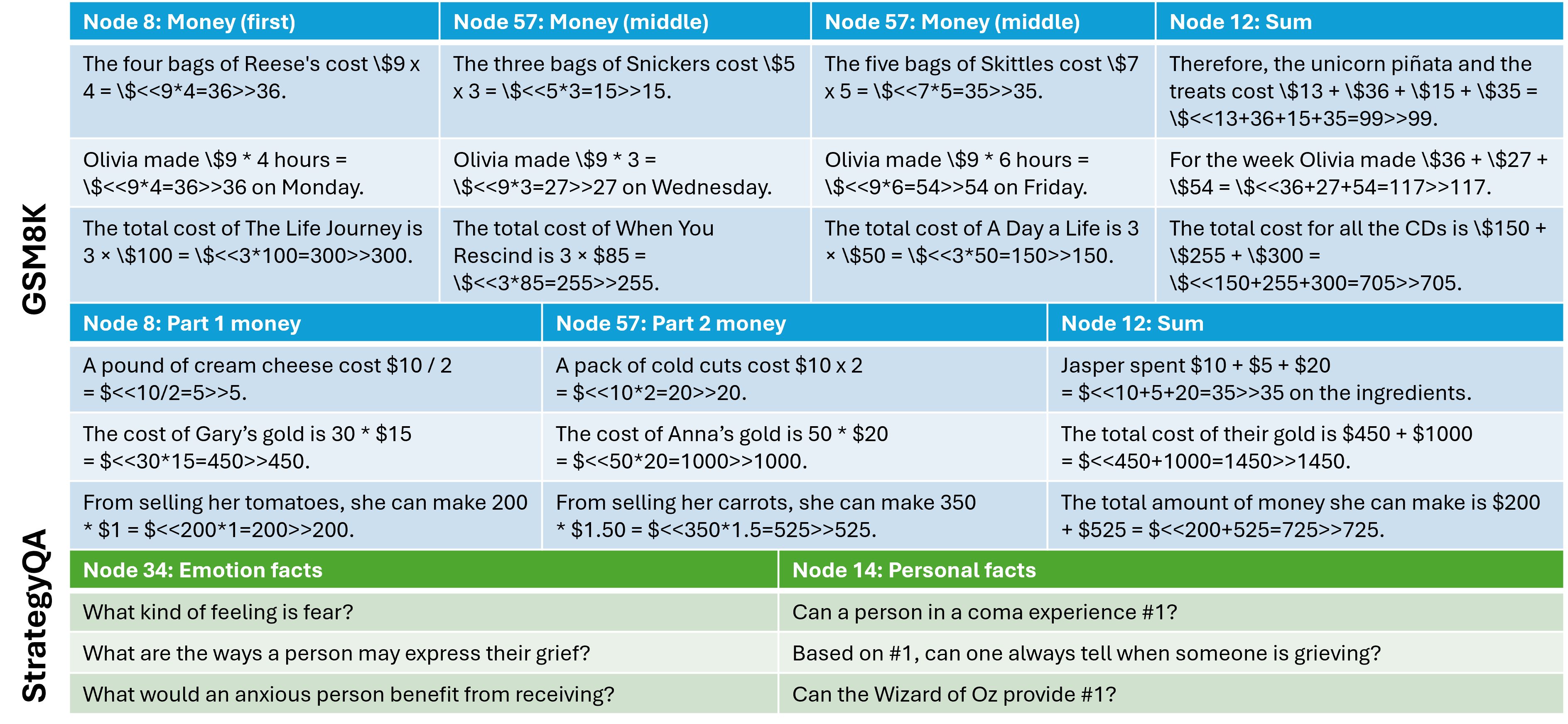}
    \caption{Additional high-frequency node patterns in the training data of \textbf{GSM8K} and \textbf{StrategyQA}, discovered by our constructed latent reasoning graphs, with example CoT solutions belonging to the node pattern.}
    \label{fig:cot_example_app}
\end{figure*}

With the \textbf{GSM8K} examples in \cref{fig:cot_example_app}, we show that our method discovers a pattern that first computes money for parallel items/individuals and then sums them up, within 3 and 4 steps respectively. With the \textbf{StrategyQA} example in \cref{fig:cot_example_app}, we show a 2-step pattern that first asks about an emotion/psychology fact and then asks the applicability to an individual in the second question.

\section{Limitations}
While the scope of this project is to provide a plausible understanding of how language models obtain reasoning abilities from next-token pre-training, we acknowledge that there are other possible ways of understanding this phenomenon. While our empirical results show our hypothesis is also effective in real-world reasoning tasks, our experiments remain on a small scale with specific tasks, limited by our computation resources and project scope. An important future work is to apply our proposed random walk training method to a large and diverse reasoning corpus with more training steps in the actual pre-training phase and verify the effectiveness of our method in improving the general reasoning ability of LLMs. We also want to point out that our proposed method is effectively up-sampling the given training set and might amplify unwanted artifacts/biases if exist in the original dataset.

\end{document}